\documentclass[runningheads]{llncs}
\usepackage[T1]{fontenc}
\usepackage{graphicx}
\usepackage{lipsum}
\usepackage{amsmath,amssymb}
\usepackage{caption}
\usepackage{subcaption}
\usepackage{tabularray}
\usepackage{algorithm,algorithmic}
\begin{document}
\title{Regularizing CNNs using Confusion Penalty Based Label Smoothing for Histopathology Images}
%
%
\author{Somenath Kuiry\inst{1}\orcidID{0000-0002-8462-5547} \and
Alaka Das\inst{1}\orcidID{0000-0003-2023-6566} \and Mita Nasipuri\inst{2}\orcidID{0000-0002-3906-5309} \and Nibaran Das\inst{2}\orcidID{0000-0002-2426-9915}}
\authorrunning{Kuiry et al.}
%
\institute{Department of Mathematics, Jadavpur University, Kolkata-32, West Bengal, India \and Department of CSE, Jadavpur University, Kolkata-32, West Bengal, India 
\\
\email{mitanasipuri@gmail.com}\\
\email{\{skuiry.math.rs, alaka.das, nibaran.das\}@jadavpuruniversity.in}}
\maketitle              
\begin{abstract}
Deep Learning, particularly Convolutional Neural Networks (CNN), has been successful in computer vision tasks and medical image analysis. However, modern CNNs can be overconfident, making them difficult to deploy in real-world scenarios. Researchers propose regularizing techniques, such as Label Smoothing (LS), which introduces soft labels for training data, making the classifier more regularized. LS captures disagreements or lack of confidence in the training phase, making the classifier more regularized. Although LS is quite simple and effective, traditional LS techniques utilize a weighted average between target distribution and a uniform distribution across the classes, which limits the objective of LS as well as the performance. This paper introduces a novel LS technique based on the confusion penalty, which treats model confusion for each class with more importance than others. We have performed extensive experiments with well-known CNN architectures with this technique on publicly available Colorectal Histology datasets and got satisfactory results. Also, we have compared our findings with the State-of-the-art and shown our method's efficacy with Reliability diagrams and t-distributed Stochastic Neighbor Embedding (t-SNE) plots of feature space.

\end{abstract}
\section{Introduction} \label{Introduction}
\begin{figure}
     \centering
     \begin{subfigure}[b]{0.4\textwidth}
         \centering
         \includegraphics[width=\textwidth]{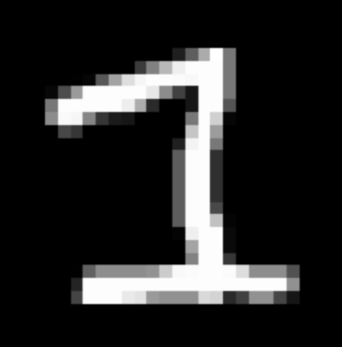}
         \caption{Example Image}
         \label{sample}
     \end{subfigure}  
     \hfill
     \begin{subfigure}[b]{0.4\textwidth}
         \centering
         \includegraphics[width=\textwidth]{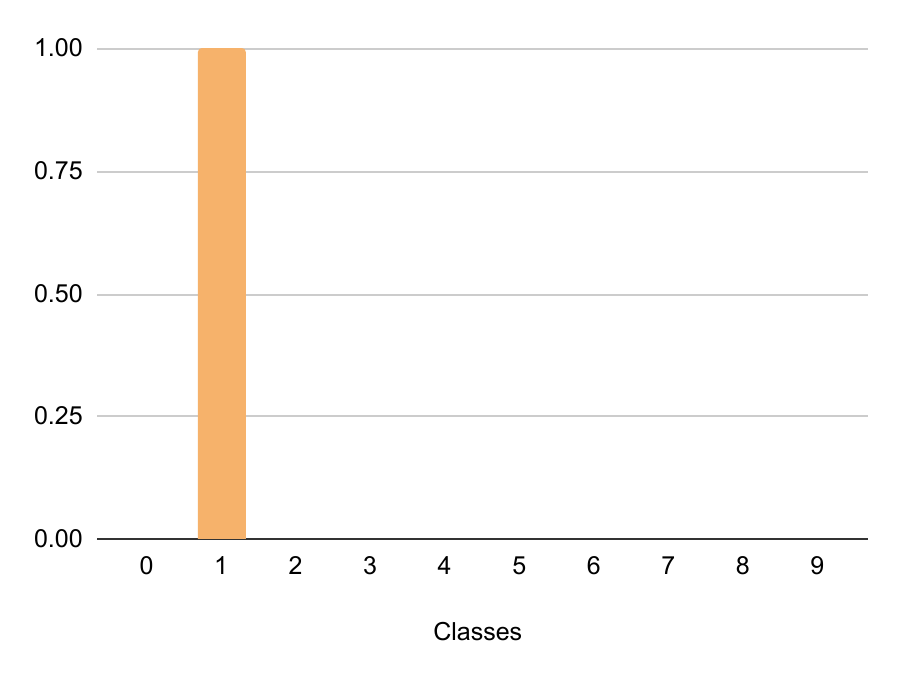}
         \caption{Hard Label}
         \label{hard}
     \end{subfigure}
  
     \begin{subfigure}[b]{0.4\textwidth}
         \centering
         \includegraphics[width=\textwidth]{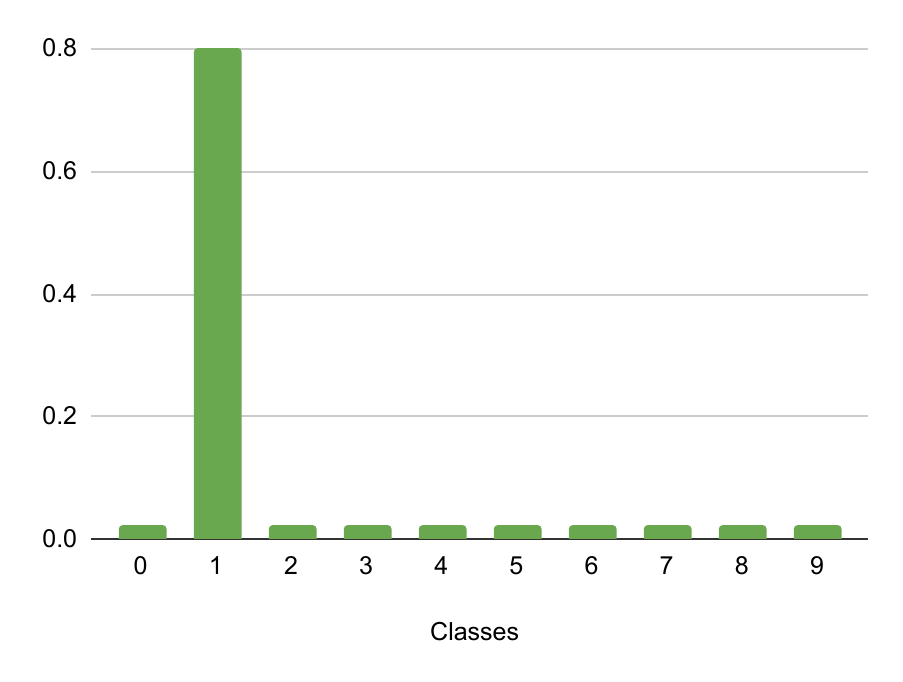}
         \caption{Label Smoothing\cite{szegedy2016rethinking}}
         \label{soft}
     \end{subfigure}
     \hfill
     \begin{subfigure}[b]{0.4\textwidth}
         \centering
         \includegraphics[width=\textwidth]{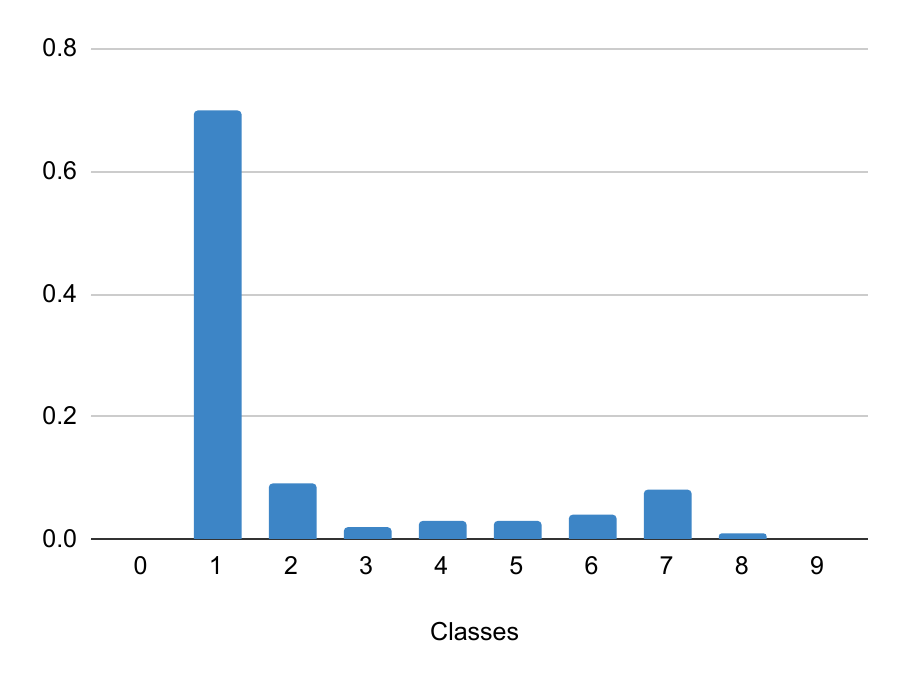}
         \caption{Ours}
         \label{our}
     \end{subfigure}
        \caption{This is an example image taken from the MNIST dataset with the correct label as "1". The differences between Hard label, Vanilla label smoothing, and our label smoothing technique. Since the image has features of class "7" and "2", the weightage is given to those classes more than others, unlike the vanilla smoothing technique.  }
        \label{total figure}
\end{figure}

Deep Learning especially Convolutional Neural Networks(CNN) has shown remarkable success in various computer vision tasks in past decades and has been adopted in medical image analysis as well \cite{shen2017deep}\cite{litjens2017survey}\cite{razzak2018deep}. However, modern CNNs tend to be overconfident about their predictions making them hard to deploy in real-world scenarios \cite{wang2021rethinking}. Researchers proposed various regularizing techniques \cite{wei2022mitigating}, \cite{reed2014training}, \cite{devries2017improved} \cite{zhang2017mixup} \cite{ghiasi2018dropblock} etc. and Label Smoothing(LS) \cite{szegedy2016rethinking} is one of them. Instead of using hard labels during training, the LS introduces soft labels for training data making the training classifier more regularized. In this process, a small amount of weight is taken from the hard label target class and redistributed with the other classes equally(see Fig.\ref{soft}).
Another way of viewing the effectiveness of LS is in its application in medical image analysis. In benchmark datasets like CIFAR-10, ImageNet, MNIST, etc., the annotations are well-defined and have a high annotator agreement, which is not the case for the medical domain. There is a high annotator disagreement present between different annotators in the medical field. For example, in most situations, two radiologists cannot say with 100\% confidence whether a particular image of a tumor is cancerous or not. Hence, using hard labels can make a classifier overconfident about its predictions. Thus LS can capture the disagreement or lack of confidence in the training phase making the classifier more regularized.
Although the vanilla LS is simple and works quite well, this limits the objective of label smoothing as it treats all the other classes with equal importance \cite{zhang2021delving}. In this paper, we have introduced a novel LS technique based on the confusion penalty. Basically, in every epoch, we keep noting the model confusion for each class. Instead of giving equal weightage to the rest of the classes, we gave weightage to those classes with whom model confusion is higher(see Fig. \ref{our}). In summary, the contributions are as follows
\begin{itemize}
    \item We have introduced a novel Label Smoothing technique for model regularization based on confusion penalty.
    \item We have performed extensive experiments with well-known classifier models and got satisfactory performance. 
    \item We have shown the feature space t-distributed Stochastic Neighbor Embedding(t-SNE) visualizations to establish the effectiveness
    \item The proposed method has been compared with other State-of-the-art techniques with the publicly available Colorectal histology dataset\cite{kather2016multi}.
\end{itemize}
The rest of the paper is organized as follows. In section \ref{previous}, we briefly discussed the previous works about LS. In section \ref{methodology}, we discussed the whole methodology in detail. The experiments and findings of these experiments are shown in section \ref{experiment} and section \ref{results} respectively. Finally, we gave the conclusion in section \ref{conclusion}

\section{Related works} \label{previous}
Training with hard labels leads to overconfident models which is not appropriate for tasks with high risk factors. Label smoothing \cite{szegedy2016rethinking} was first proposed in 2016, as a regularization technique to reduce the model overconfidence. Though the method is simple and intuitive, the drawback of this is it treats all other classes except the target class equally and assigns equal weights for them. 

Pham et al. \cite{21} suggested remapping targets in the field of medical image analysis to random values that are near to 1. They discovered that this approach increased model performance on the CheXpert Dataset \cite{13} by about 1.4\%. Additionally, Xi et al. \cite{18} addressed accuracy issues by utilizing a spatial label smoothing technique to achieve sufficient performance with less reliance on well-annotated data. Krothapalli and Abbott\cite{Als} proposed an adaptive label smoothing based on relative object size within an image. For context-only images, their method penalizes low-entropy (high-confidence) predictions while simultaneously assisting in the generation of accurate labels during training.

More recently, Wei et al. \cite{hist} utilized agreement-aware and confidence-aware label smoothing for calibrating neural networks for histopathology images. Zhang et al. \cite{zhang2021delving}, introduced a novel label smoothing technique Online label smoothing(OLS) on the CIFAR-100 dataset with ResNet-101 architecture.

\section{Methodology} \label{methodology}
\subsection{Preliminaries} \label{preli}
Let $\mathcal{D} = \{(x_i,y_i)\}$, be the dataset where $x_i$ and $y_i$ denote the images and targets respectively. Let $p$ and $\theta$ be the prediction probability distribution of the model and the target class distribution for a particular sample $(x_i,y_i)$. In the traditional training with hard labels, a one-hot encoding($1$ for the correct class and $0$ for the incorrect classes) is used as the target distribution. So, for hard labels, we have $\theta(c == y_i | x_i) = 1$ and $\theta(c \neq y_i | x_i) = 0$ for $c = 1,2, \dots C$, where $C$ is the total number of classes. Also, $p(c|x_i)$ is the probability score of input $x_i$ for the class $c$. Hence, the traditional cross-entropy loss function for that sample $(x_i, y_i)$ would be 
\begin{align}
 \mathcal{L}_{\text{Hard}} & = -\sum_{c = 1}^{C} \theta(c|x_i) \log p(c|x_i) \nonumber \\
                           & = - \log p(c == y_i|x_i)   
\end{align}
In the case of a vanilla soft label, we just replace the target distribution $\theta$ with a smooth target distribution $\phi$ by a weighted ($\alpha$) average between the target distribution $\theta$ and a uniform distribution across the classes. The parameter $\alpha$ is called a smoothing parameter and the cross entropy loss function for this case is derived as
\begin{align}
    \mathcal{L}_{LS} = -\sum_{c = 1}^{C} \left[(1-\alpha)*\theta(c|x_i) + \dfrac{\alpha}{C} \right] \log p(c|x_i)
\end{align}

\subsection{Confusion Penalty based Label Smoothing(CPLS)}

\begin{figure}
    \centering
    \includegraphics[height= 0.8\linewidth,width=\textwidth]{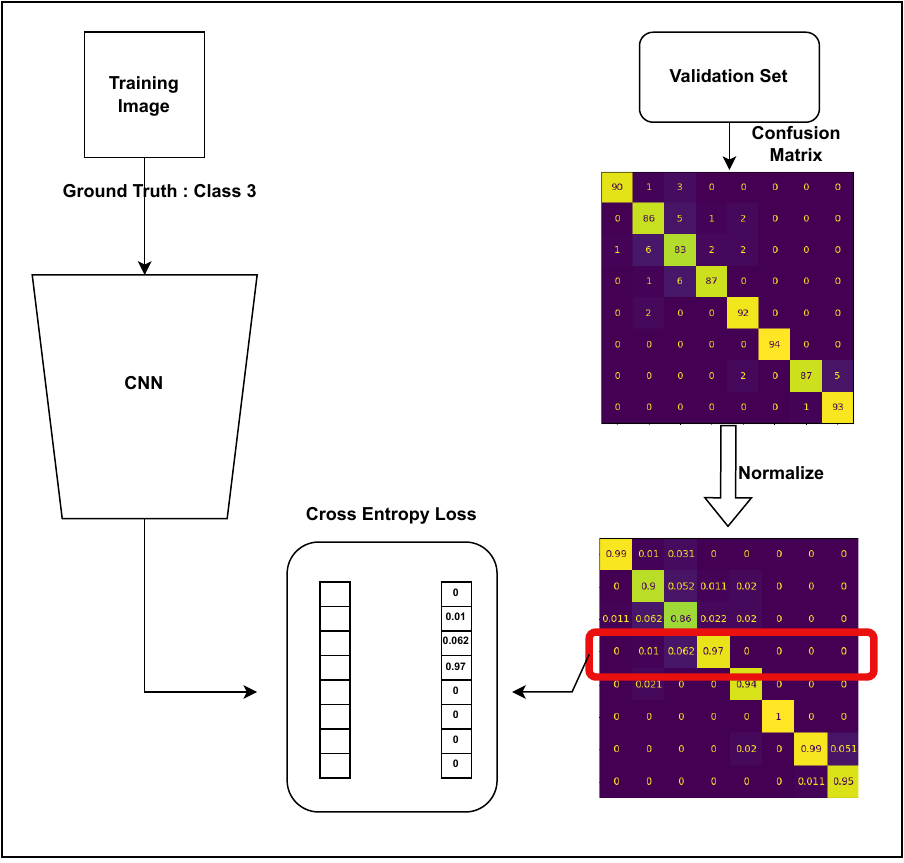}
    \caption{The training procedure of our CPLS method.}
    \label{main}
\end{figure}

Though vanilla label smoothing works quite well in practice, it gives equal importance to all the other classes and overlooks the fact that not every class has similar characteristics \cite{zhang2021delving}. In this paper, we introduce the Confusion Penalty Label Smoothing (CPLS), which captures the relationship or similarity between classes. The key idea here is to distribute the probabilities among the classes with similar features according to feature similarity. To find out these intra-class similarities, we keep track of the images from a particular class that is often confused with images belonging to other classes. This is done by simply saving the confusion matrix of the validation data. 
Let $M_n=(m_{ij})$ denote the confusion matrix of validation data in epoch number $n$, where $n = 1, 2...$ etc. $M_n$ is a $C \times C$ matrix where the rows denote the true class and the columns denote the predicted classes. Thus each cell $j$ of row $i$ of $M_n$ indicates out of the total sample from class $i$, how many sample is classified as class $j$. To use this information as a target distribution, we normalize each confusion matrix row-wise, which will now indicate the confusion distribution of each class. We then take this distribution and use this in the next epoch as a smooth label using the following loss function
\begin{align}
    \mathcal{L}_{\text{CPLS}} = -\sum_{c = 1}^{C} m_{ic}\log p(c|x_i) 
\end{align}
The training procedure is shown in the Figure \ref{main}. In the next section, we discussed the detailed experimental setup of this work.

\section{Experiments} \label{experiment}

 \begin{algorithm}[t]
 \footnotesize

 \caption{Training Procedure with CPLS }
 \begin{algorithmic}[1]
 \renewcommand{\algorithmicrequire}{\textbf{Constants:}}

 \REQUIRE  
    $\mathcal{D}_{train}=\{x_i,y_i\}$ = Training data with labels \\
	$\mathcal{D}_{val}=\{x_i,y_i\}$ = Validation data with labels \\
	$p(c|x_i)$ = Softmax output of image $x_i$ for class $c$\\
	$M_n=(m_{ij})$ = Confusion Matrix of validation set. Initialize with Identity matrix\\
	N = Threshold \\
        $0 < \beta <1$

 \ENSURE

  \IF{EPOCH $\leq$ N}
    \STATE
          \FOR {$x_i$ to $\mathcal{D}_{train}$}
          \STATE
                $\mathcal{L}_{\text{Hard}} = - \log p(c == y_i|x_i)$  \\
                Loss = $\mathcal{L}_{\text{Hard}}$
         \ENDFOR
         \RETURN Loss
  \ELSIF{EPOCH $>$ N}
     \STATE
          \FOR {$x_i$ to $\mathcal{D}_{train}$}
          \STATE
                $\mathcal{L}_{\text{CPLS}} = -\sum_{c = 1}^{C} m_{ic}\log p(c|x_i)$   \\
                Loss = $\beta\mathcal{L}_{\text{Hard}} + (1-\beta)\mathcal{L}_{\text{CPLS}}$
         \ENDFOR 
         \FOR {$x_i$ to $\mathcal{D}_{val}$}
          \STATE
                Calculate Confusion matrix $M_n$ \\
                $M_n = \operatorname{Normalized} (M_n)$
                
         \ENDFOR 
         \RETURN Loss
 \ENDIF
 \end{algorithmic}
 \end{algorithm}

\subsection{Dataset}
In this work, we have considered the publicly available  Colorectal$\_$histology \cite{kather2016multi}. This dataset consists of $5000$ histological images of human colorectal cancer. Each image has dimensions of $150 \times 150$ pixels and belongs to one of eight classes: `TUMOR', `STROMA', `COMPLEX', `LYMPHO', `DEBRIS', `MUCOSA', `ADIPOSE', and `EMPTY'. This dataset is well-balanced, with an equal number of samples for each class. For training, validation, and testing, we divided the data in a $70:15:15$ ratio, respectively. 
\subsection{Experimental Setup}
\begin{figure}
     \centering
     \begin{subfigure}[b]{0.49\textwidth}
         \centering
         \includegraphics[width=\textwidth]{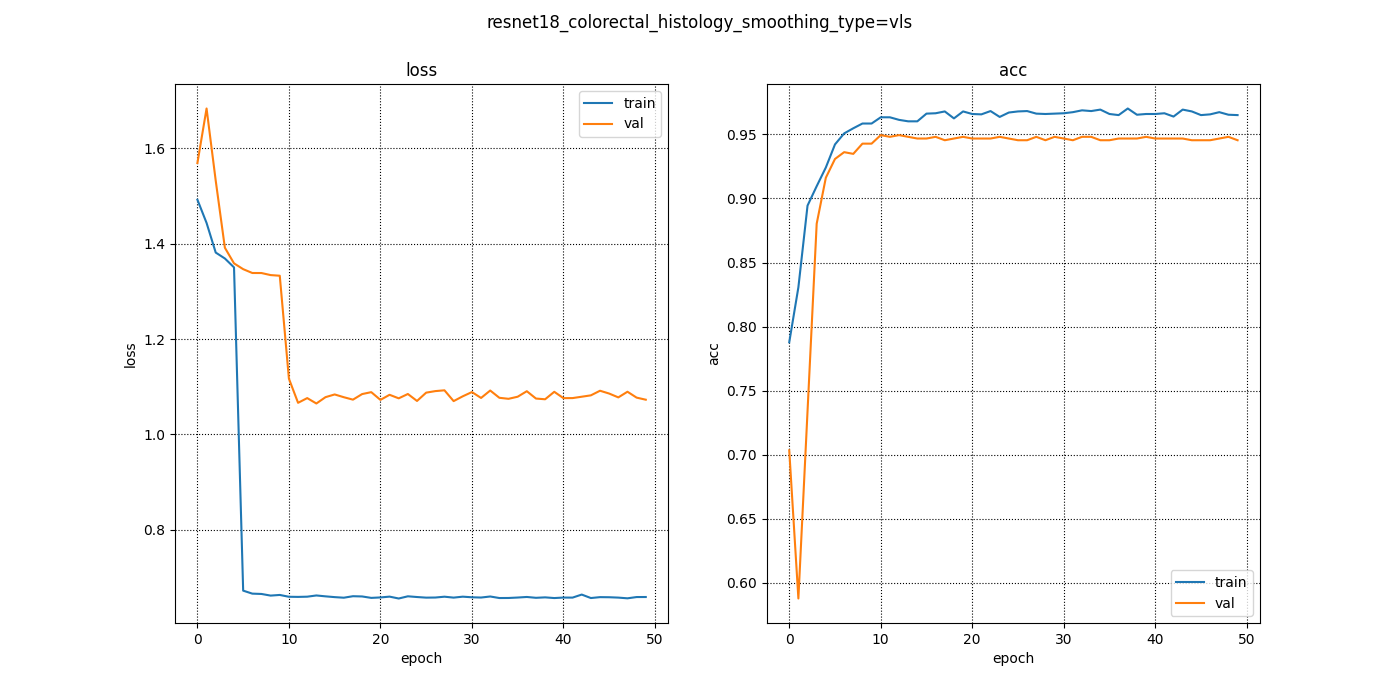}
         \caption{ResNet-18}
     \end{subfigure}  
     \hfill
     \begin{subfigure}[b]{0.49\textwidth}
         \centering
         \includegraphics[width=\textwidth]{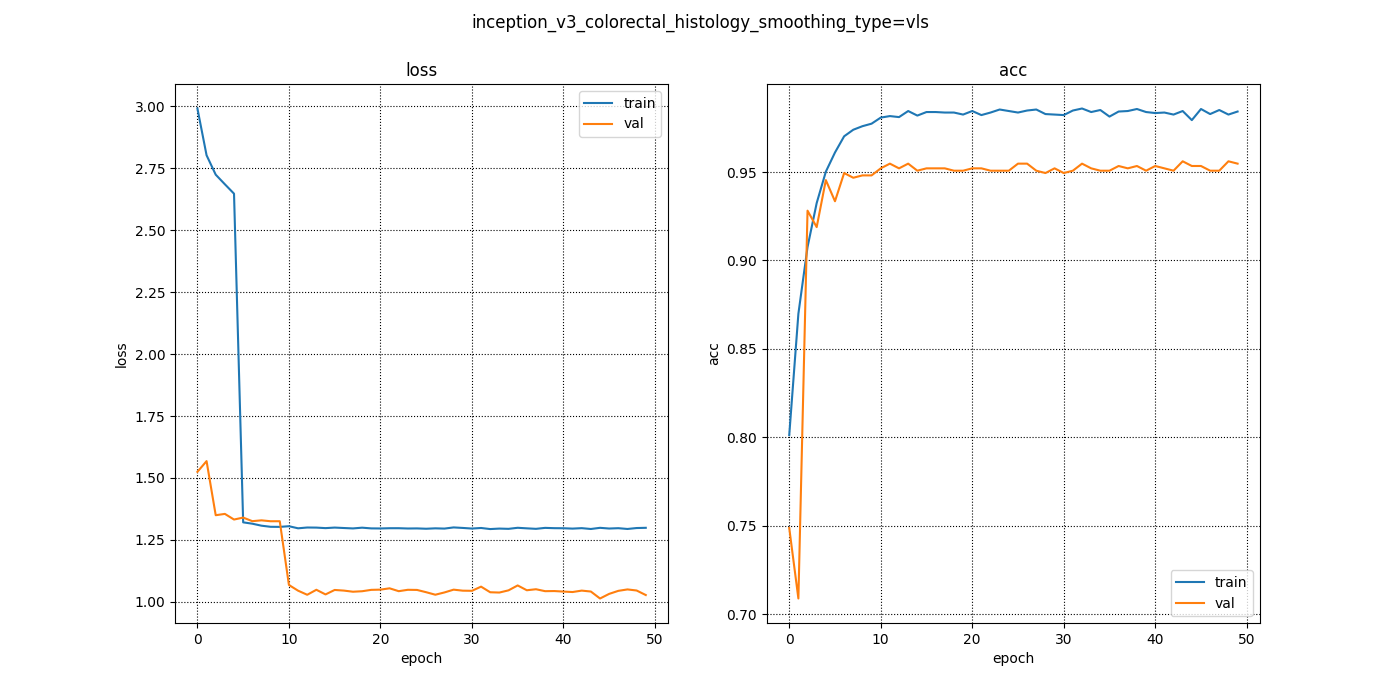}
         \caption{Inception v3}
     \end{subfigure}
  
     \begin{subfigure}[b]{0.49\textwidth}
         \centering
         \includegraphics[width=\textwidth]{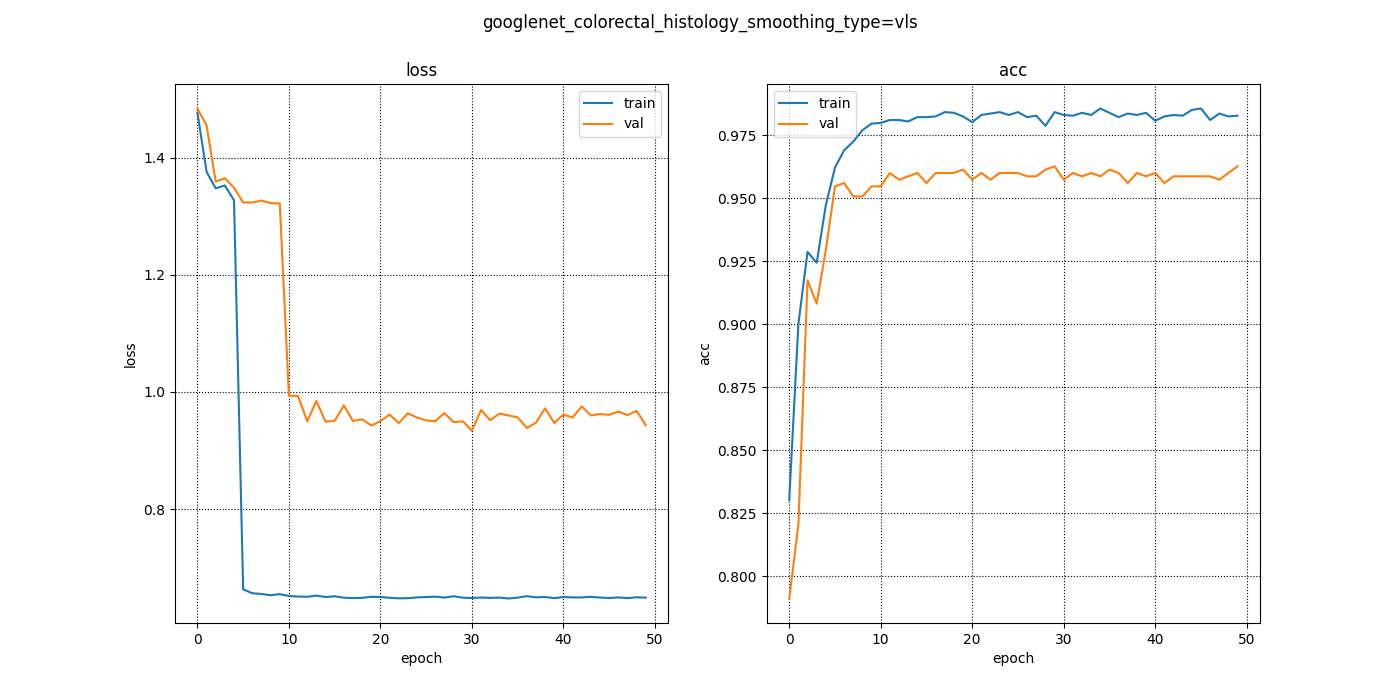}
         \caption{GoogLeNet}
     \end{subfigure}
     \hfill
     \begin{subfigure}[b]{0.49\textwidth}
         \centering
         \includegraphics[width=\textwidth]{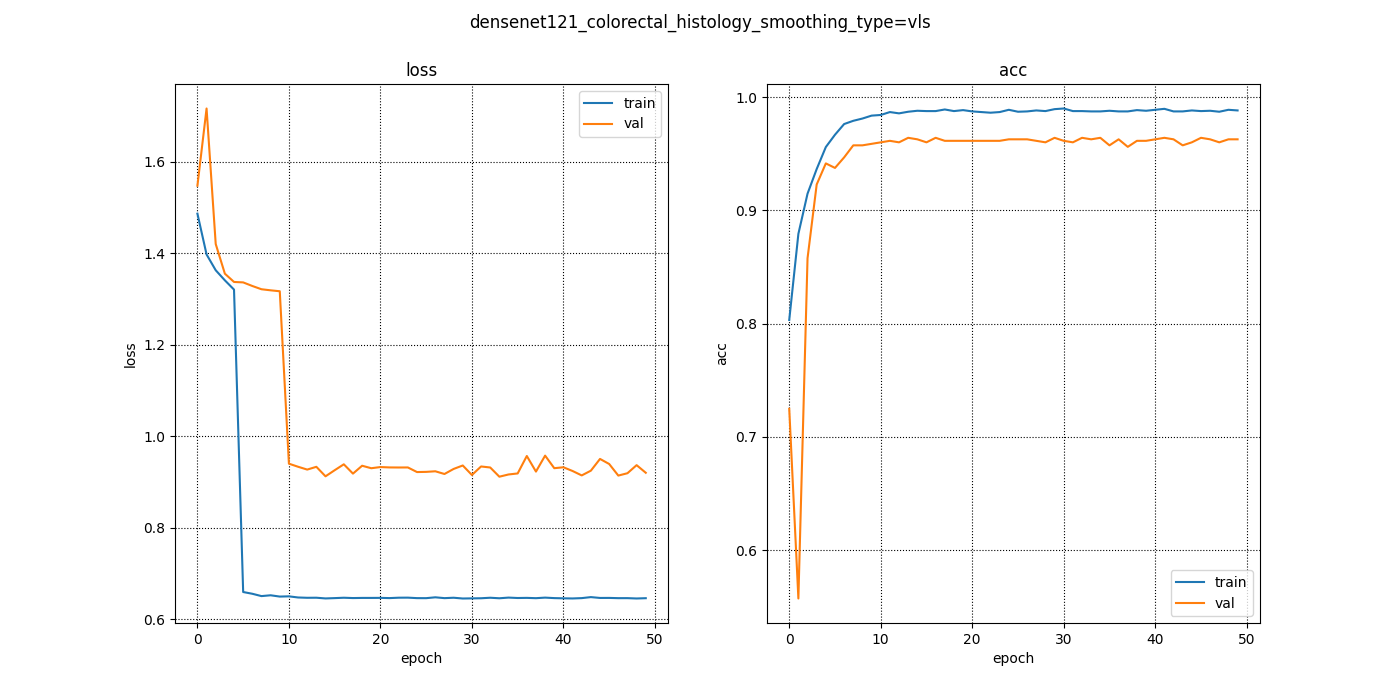}
         \caption{DenseNet-121}
     \end{subfigure}
     \begin{subfigure}[b]{0.5\textwidth}
         \centering
         \includegraphics[width=\textwidth]{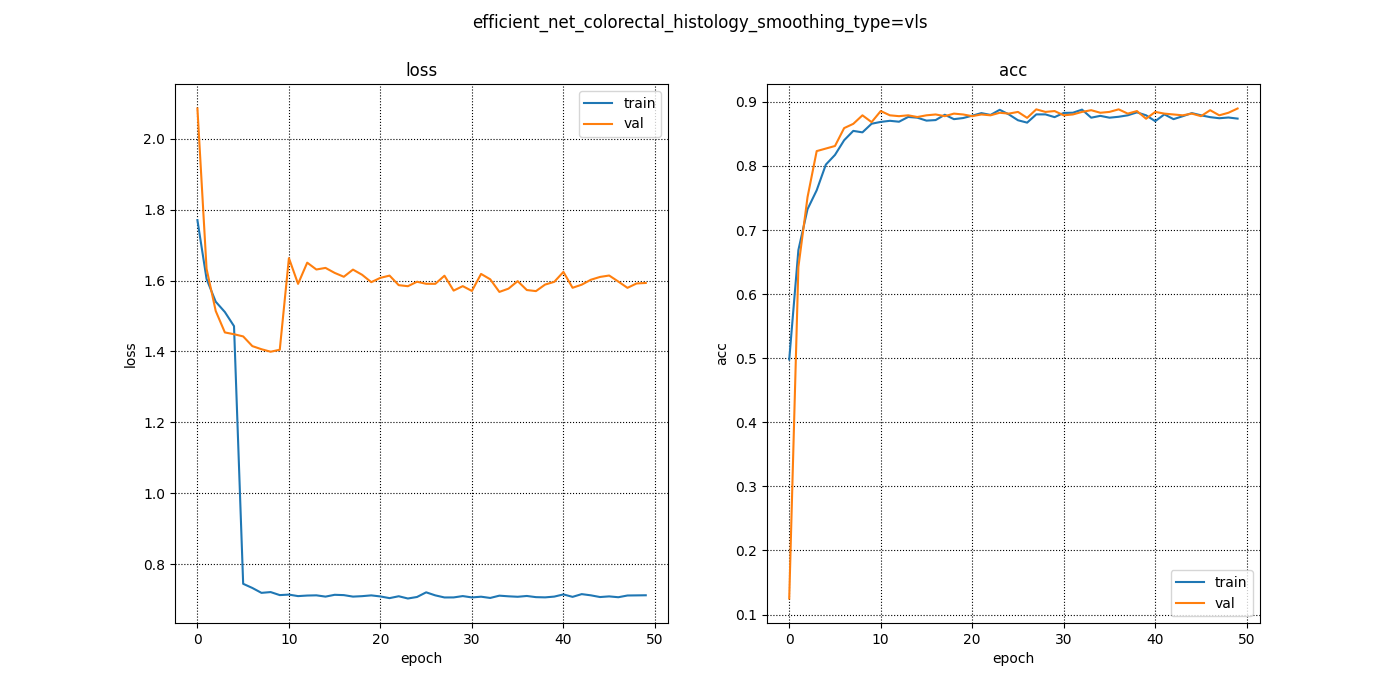}
         \caption{EfficientNet}
     \end{subfigure}
        \caption{Loss and Accuracy curves of all the classifiers. Blue and Orange are for training and validation sets respectively}
        \label{loss}
\end{figure}

We implemented various state-of-the-art CNN classifiers available in the PyTorch library, including DenseNet-121\cite{DenseNet}, GoogLeNet\cite{googlenet}, ResNet-18\cite{resnet}, Inception V3\cite{inception}, and EfficientNet \cite{efficient}. These classifiers were trained from scratch with similar hyperparameters, which made the experiment protocol simple and minimalistic. We did not use any kind of data augmentation as well to see the effectiveness of the label smoothing.  We take these three models for further experiments with label smoothing. For the loss function, we have used the $\mathcal{L}_{\text{CPLS}}$ loss initially. The problem that we faced here is that in the starting phase, the model did not have any information about the confusion matrix, and with a random initialization, the model's performance degrades as it lacks the hard label from the beginning and the model tends to diverge. Hence, we have employed a hybrid loss function $\mathcal{L}$ as described in Equation \ref{eq4}, and the whole training process is given in Algorithm 1.
\begin{align}
    \mathcal{L} = \beta \mathcal{L}_{\textbf{Hard}} + (1-\beta)  \mathcal{L}_{\textbf{CPLS}} \label{eq4}
\end{align}
During training, we employ the hard label loss $\mathcal{L}_{\textbf{Hard}}$ for a few epochs then we introduce the new loss function $\mathcal{L}$ into the training. The advantage of this is that initially, the model gets more and more confident about the images for a few epochs and then the new loss function keeps the model from over-fitting. Also, the model can have a more accurate confusion matrix for the updated loss to work with.
Each classifier was trained with the same hyperparameters to make the experiment protocol simple. These models were trained for around $50$ epochs with NVIDIA GeForce Quadro P5000
$16$ GB GPU in PyTorch environment. The loss curves are shown in the Figure \ref{loss}.    

\section{Result and Discussion} \label{results}
A well-calibrated classifier's prediction probability reflects the true likelihood of the event of interest. For example, if a classifier predicts $N$ number of images with 
the highest probability equal to $0.8$, then $80\%$ of those $N$ images should be correct. The Expected Calibration Error(ECE)\cite{ece} measures this model calibration by calculating the weighted average error between the prediction confidence(probabilities) and accurate prediction percentage. In practice, we divide the confidence range(0 to 1) into a few bins(n) and calculate the weighted average of the differences between the accurate prediction percentage and the prediction confidence. The mathematical expression is given as
$$
\text{ECE} = \sum_{m=1}^{n} \dfrac{|B_m|}{n} |\operatorname{acc}(B_m) - \operatorname{conf}(B_m)|
$$
Here, $B_m$ is a particular bin, $acc(B_m)$ is the percentage of correct classification, and $conf(B_m)$ is the confidence of those samples. The lower ECE indicates a well-calibrated model and is suitable for application. In this paper, we have considered the overall testing accuracy and Expectation Calibration Error(ECE)\cite{ece} as the evaluation metric for our primary classifier ResNet-18, Inception V3, EffienetNet V2, GooLeNet, and DenseNet-121.

\begin{table}
\centering
\caption{Comparision of Testing Accuracy and ECE with Hard label, vanilla Soft label\cite{szegedy2016rethinking}, Online label smoothing\cite{zhang2021delving}, and our CPLS method. Here the terms hard, vanilla, and ols represent Hard label, Vanilla Soft label, and Online label smoothing respectively.}
\begin{tblr}{
  width = \linewidth,
  colspec = {Q[529]Q[237]Q[142]},
  column{2} = {c},
  column{3} = {c},
  hline{1-2,6,10,14,18,22} = {-}{},
}
\textbf{Networks} & \textbf{Accuracy} & \textbf{ECE}\\
Resnet 18 + hard& 0.9521 & 2.75\\
Resnet 18 + vanilla & 0.9534 & 2.69\\
Resnet 18 + ols & 0.9521 & 2.76\\
Resnet 18 + cpls & \textbf{0.9601} & \textbf{1.75}\\
InceptionV3 + hard & 0.9627 & 2.77\\
InceptionV3 + vanilla  & 0.9654 & 2.56\\
InceptionV3 + ols & 0.9601 & 2.48\\
InceptionV3 + cpls & \textbf{0.9694} & \textbf{2.42}\\
DenseNet 121 + hard & 0.9694 & 2.52\\
DenseNet 121 + vanilla  & 0.9654 & 2.5\\
DenseNet 121 + ols & 0.964 & 2.55\\
DenseNet 121 + cpls & \textbf{0.9734} & \textbf{1.77}\\
GoogLeNet + hard & 0.964 & 2.75\\
GoogLeNet + vanilla & 0.972 & 2.61\\
GoogLeNet + ols & \textbf{0.98} & \textbf{1.89}\\
GoogLeNet + cpls & 0.964 & 2.13\\
EfficientNet + hard & \textbf{0.8989} & 12.86\\
EfficientNet + vanilla & 0.851 & \textbf{8.62}\\
EfficientNet + ols & 0.8789 & 9.09\\
EfficientNet + cpls & 0.8896 & 8.81
\end{tblr}
 
\label{res}
\end{table}

\begin{figure}
    \centering
    \includegraphics[height=1.86\linewidth]{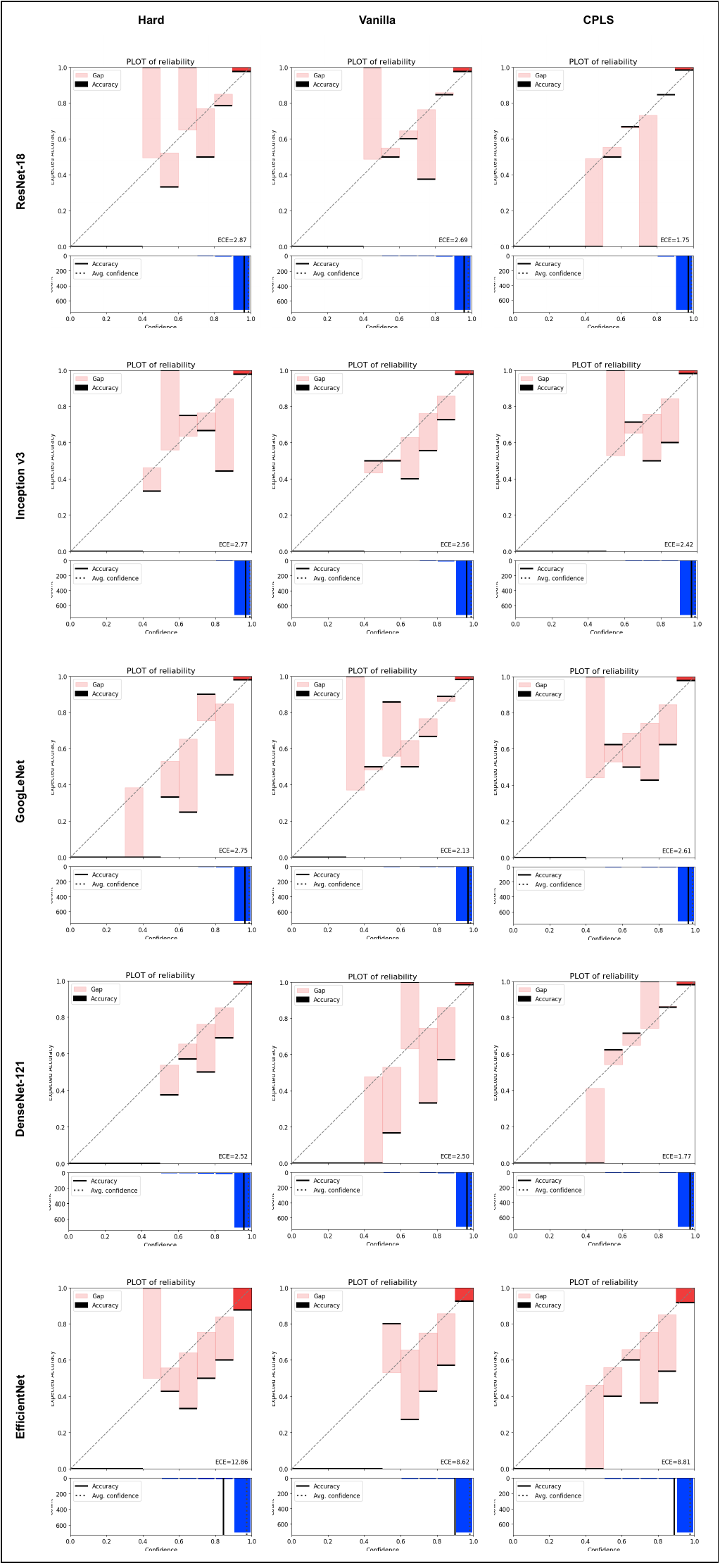}
    \caption{Comparison of Reliability Diagrams between all the classifiers with hard label, vanilla label smoothing, and our technique. }
    \label{rd}
\end{figure}
In Table \ref{res}, we have shown the Testing accuracy and ECE obtained by different classifiers. Each network is trained with State-of-the-art smoothing techniques like vanilla label smoothing, and online label smoothing to compare with ours. 
From Table \ref{res}, we can see that vanilla and online label smoothing perform better than training with hard labels for almost every classifier apart from Efficient Net in terms of Test accuracy and ECE. However, our method CPLS outperforms every other in most situations. For GoogLeNet, Online label smoothing has the best performance. For EfficinetNet, hard labels achieve the best test accuracy but the vanilla label smoothing has the lowest ECE.
In Figure \ref{rd}, we have presented the Reliability Diagrams of each network trained with Hard label, Vanilla soft label, and our CPLS. These diagrams correspond to the ECE scores of each network. 

\begin{figure}
    \centering
    \includegraphics[width=\textwidth]{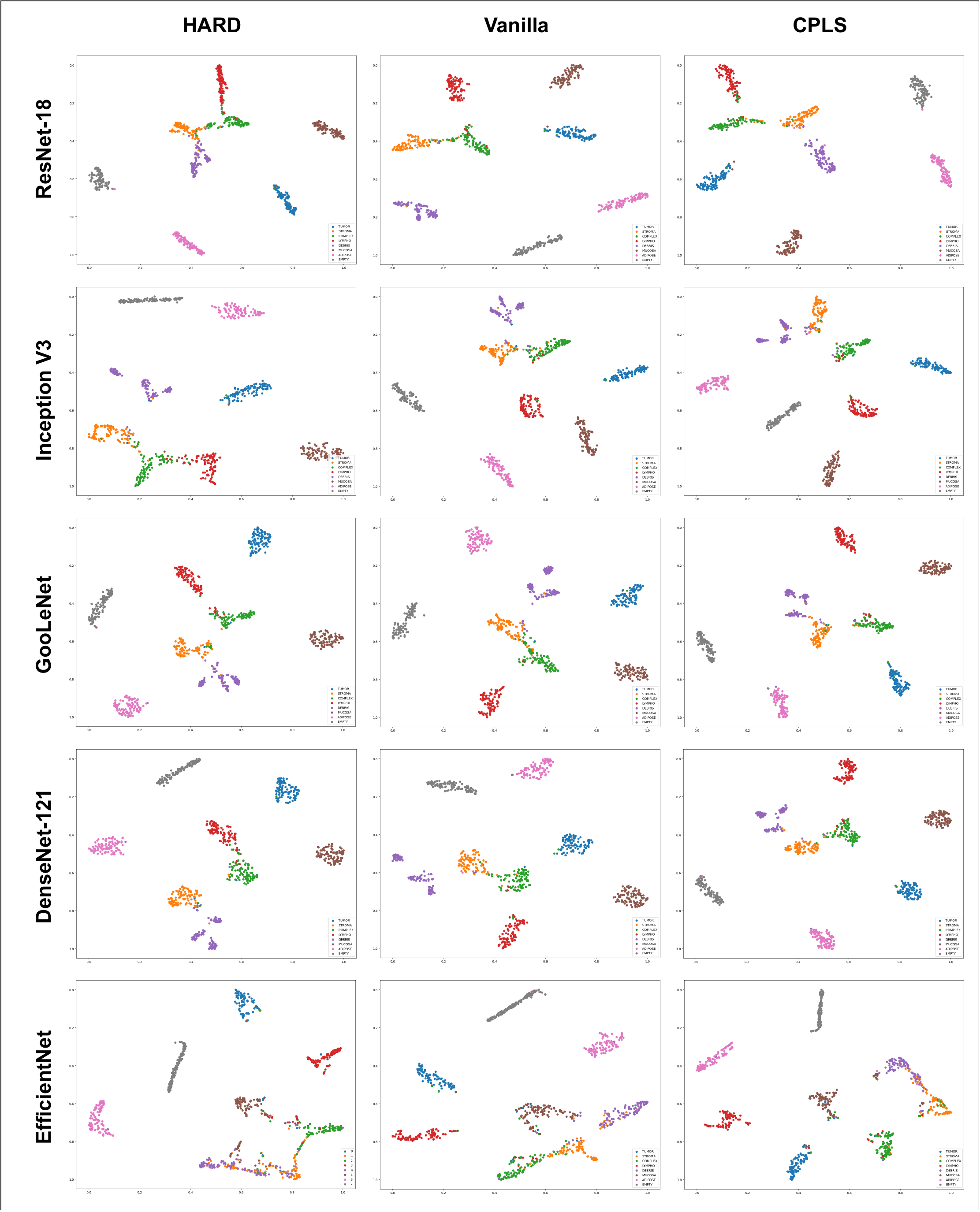}
    \caption{t-SNE plots of feature space for all the classifiers trained with the hard label, vanilla label smoothing, and our technique. }
    \label{tsne}
\end{figure}

For further analysis, we have shown the TSNE plots of feature space of each network trained with hard label, vanilla soft label, and our CPLS in Figure \ref{tsne}. In almost all situations, vanilla soft label created better clusters than the hard label. However, our CPLS techniques created better clusters in the feature space. This indicates the efficacy of our novel CPLS technique.

\section{Conclusion} \label{conclusion}
In this paper, we proposed a novel label smoothing techniques for model regularization. To create such a smooth label from a hard label, we utilized the confusion between frequent classes as the primary information from the validation dataset. Unlike the vanilla label smoothing, our technique does not give importance to each class equally. ResNet-18, Inception V3, GooLeNet, and EfficientNet were used in this experiment on the Colorectal Histology dataset. The Expected Calibration Error(ECE) is 1.35\%, 0.5\%, and 0.61\% less on average than the hard label, vanilla label smoothing, and Online Label smoothing respectively. From the TSNE plots (Figure \ref{tsne}) of feature space, we can see that our CPLS can create better clusters than hard labels and vanilla soft labels, which proves the effectiveness of our novel label smoothing technique. Though in terms of accuracy and ECE, our method works well, it makes the models under-confident in some cases which can be seen in the Reliability Diagrams(Figure \ref{rd}). In the future, we would like to address this issue. Also, we would like to use this technique in the domain of image segmentation as well.

\begin{credits}
\subsubsection{\ackname} This research was carried out at the Centre for Microprocessor Application for Training Education and Research Lab within the Computer Science and Engineering Department of Jadavpur University and received partial support from a project funded by SERB, Govt. of India,(No: EEQ/2018/000963) and DST, GOI through the INSPIRE Fellowship program (IF170641).

\subsubsection{\discintname}
The authors have no conflict of interest to disclose.
\end{credits}
%
%
%
%




\bibliographystyle{plain}
\bibliography{ref.bib}

\end{document}